\documentclass[journal]{IEEEtran}
\usepackage{graphicx, times, amsmath, amssymb, cite, subfigure, subfigure,booktabs, mathtools,bm,threeparttable,multirow,url}
\usepackage[ruled,norelsize]{algorithm2e}

\allowdisplaybreaks

\makeatletter
\newcommand{\removelatexerror}{\let\@latex@error\@gobble}
\makeatother
\begin{document}

\title{Supervised Enhanced Soft Subspace Clustering (SESSC) for TSK Fuzzy Classifiers}
\author{
    Yuqi~Cui, Huidong~Wang and  Dongrui~Wu
    \thanks{
        Y.~Cui and D.~Wu are with the Key Laboratory of the Ministry of Education for Image Processing and Intelligent Control, School of Artificial Intelligence and Automation, Huazhong University of Science and Technology, Wuhan 430074, China. Email: yqcui@hust.edu.cn, drwu@hust.edu.cn.}
    \thanks{
        H.~Wang is with the School of Management Science and Engineering, Shandong University of Finance and Economics, Jinan 250014, China. Email: huidong.wang@ia.ac.cn.}
    \thanks{
        Dongrui~Wu is the corresponding author.
    }
}

\maketitle

\begin{abstract}
Fuzzy $c$-means based clustering algorithms are frequently used for Takagi-Sugeno-Kang (TSK) fuzzy classifier antecedent parameter estimation. One rule is initialized from each cluster. However, most of these clustering algorithms are unsupervised, which waste valuable label information in the training data. This paper proposes a supervised enhanced soft subspace clustering (SESSC) algorithm, which considers simultaneously the within-cluster compactness, between-cluster separation, and label information in clustering. It can effectively deal with high-dimensional data, be used as a classifier alone, or be integrated into a TSK fuzzy classifier to further improve its performance. Experiments on nine UCI datasets from various application domains demonstrated that SESSC based initialization outperformed other clustering approaches, especially when the number of rules is small.
\end{abstract}


\section{Introduction}

Takagi-Sugeno-Kang (TSK) fuzzy systems~\cite{nguyen2019fuzzy} have achieved great success in numerous applications~\cite{chang2008tsk, wai2004intelligent, jiang2017seizure}. They use fuzzy sets to model linguistic and numerical uncertainties~\cite{zadeh2005toward}, and IF-THEN rules to approximate the human reasoning process. Therefore, TSK fuzzy systems are more interpretable than many other (black-box) machine learning models such as neural networks. Early fuzzy systems were usually built from expert knowledge. However, data-driven modeling~\cite{rezaee2010data,drwuGD2020,cui2020optimize} has become more and more popular recently.

Many data-driven algorithms have been proposed to tune TSK fuzzy systems~\cite{bezdek1984fcm, jang1993anfis, priyono2005generation, deng2010scalable, deng2014minimax, drwuGD2020, cui2020optimize, rezaee2010data}. Optimizing a TSK fuzzy system involves fine-tuning both the antecedent parameters and the consequent parameters, which can be done separately or simultaneously~\cite{priyono2005generation,lin2006support, deng2010scalable, deng2014minimax,jang1993anfis,shi1999implementation, gacto2014metsk,drwuMTGA2020, drwuGD2020, cui2020optimize}. When they are optimized separately, the consequent parameters are usually obtained by least squares estimation (LSE), e.g., in an adaptive neuro fuzzy inference system (ANFIS)~\cite{jang1993anfis}.

However, regardless of how the antecedent and consequent parameters of a TSK fuzzy system are tuned, they must be initialized first. One of the most popular approaches for initializing the antecedents of a TSK fuzzy system is fuzzy $c$-means (FCM) clustering~\cite{bezdek1984fcm}. Unlike $k$-means clustering, in which each data sample only belongs to one cluster, FCM assigns each data sample to all clusters at different membership degrees. FCM works well on low-dimensional datasets; however, it  may fail when the data dimensionality is high~\cite{winkler2011fuzzy, deng2010enhanced} (e.g., Winkler \emph{et al.}~\cite{winkler2011fuzzy} demonstrated that randomly initialized FCM can only be used when the data dimensionality is smaller than 20), because the Euclidean distances used in it become less distinguishable.

Subspace clustering~\cite{parsons2004subspace, sim2013survey} may be used to deal with the curse of dimensionality. It selects a subset of features instead of using them all in determining the clusters. Many different subspace clustering approaches have been proposed for high dimensional datasets. They can be divided into two categories: hard subspace clustering and soft subspace clustering (SSC). The former~\cite{cheng1999entropy,goil1999mafia,aggarwal2000finding} finds exact subspaces for different clusters. On the contrary, SSC assigns weights to different features in different clusters, which reflect their contributions to the corresponding cluster. SSC may be more flexible than hard subspace clustering, and hence has drawn more attention recently~\cite{deng2016survey}.

There have been a few FCM based SSC algorithms. Keller and Klawonn~\cite{keller2000fuzzy} proposed the features weighting FCM (AWFCM), which uses weighted Euclidean distances in FCM. Frigui and Nasraoui~\cite{frigui2004unsupervised} proposed simultaneous clustering and attribute discrimination (SCAD) and its variants, which add $L_2$ regularization and an exponential term to the feature weights of AWFCM. Zhou \emph{et al.}~\cite{zhou2016fuzzy} proposed entropy weighting FCM (EWFCM) and kernel EWFCM with entropy regularization for feature weighting. Deng \emph{et al.}~\cite{deng2010enhanced} proposed enhanced SSC (ESSC) to consider the within-cluster compactness and between-cluster separation simultaneously. ESSC has also been integrated with sparse learning to construct concise TSK fuzzy systems~\cite{xu2019concise}.

All above SSC approaches are unsupervised. This paper proposes supervised ESSC (SESSC), which further takes the label information into consideration for better discriminability. SESSC can then be used to initialize TSK fuzzy classifiers. Experiments on nine real-world UCI datasets demonstrated that the proposed SESSC can indeed improve the classification performance compared with other initialization approaches, especially when the number of clusters is small.

The remainder of this paper is organized as follows: Section~\ref{sec:method} introduces background knowledge on TSK fuzzy classifiers and FCM based clustering approaches, and proposes SESSC and SESSC-LSE. Section~\ref{sec:exp} evaluates the performances of SESSC and SESSC-LSE on synthetic and real-world datasets. Section~\ref{sec:con} draws conclusions.

\section{Algorithms}\label{sec:method}

This section introduces the details of a TSK fuzzy system for multi-class classification, FCM and FCM based SSC algorithms, and our proposed SESSC and SESSC-LSE. The code for SESSC and SESSC-LSE is available at \url{https://github.com/YuqiCui/SESSC}.

The main notations are summarized in Table~\ref{tab:noti}. Matrices and vectors are denoted by uppercase letters and lowercase bold letters, respectively, e.g., $\bm{x}_n$ denotes the $i$-th row of matrix $X$.

\begin{table}[htpb]\centering
\caption{Notations used in this paper.}\label{tab:noti}
\begin{tabular} {p{50pt}p{160pt}}\toprule
Notation & Meaning  \\\midrule
      $X_{n,i}$   &   The ($n, i$)-th element of matrix $X$        \\
       $\bm{x}_n$  &    The $n$-th row of matrix $X$      \\
       $x_{n,i}$ & The $i$-th element of vector $\bm{x}_n$ \\
       $X_{\cdot,i}$ & The $i$-th column of matrix $X$ \\
       $\|\bm{x}\|_2$ & $L_2$-norm of vector $\bm{x}$\\
       $\|\bm{x}-\bm{v}\|_{\bm{w}}$& $\bm{w}$ weighted Euclidean distance between vectors $\bm{x}$ and $\bm{v}$ \\
       N & Number of training samples\\
       D& Feature dimensionality \\
       C& Number of classes\\
       R& Number of rules in a TSK fuzzy classifier \\ \bottomrule
\end{tabular}
\end{table}

\subsection{Multi-class TSK Fuzzy Classifier}\label{subsec:tsk}

Let the training dataset be $\mathcal{D}=\{\bm{x}_n, \bm{y}_n\}_{n=1}^N$, in which $\bm{x}_n=[x_{n,1},...,x_{n,D}]^T$ is a $D$-dimensional input vector, and $\bm{y}_n=[y_{n,1},...,y_{n,C}]$ the corresponding $C$-class one-hot encoding label vector. $X\in \mathbb{R}^{N \times D}$ is the input matrix containing all inputs, and $Y\in \mathbb{R}^{N\times C}$ the corresponding label matrix.

Suppose the TSK fuzzy classifier has $R$ rules, in the following form:
\begin{align}
	\begin{split}
		\textup{Rule}_r:~&\textup{IF}~x_1~\textup{is}~\tilde{X}_{r, 1}~\textup{and}~ \cdots ~\textup{and}~x_D~\textup{is}~\tilde{X}_{r, D}, \\
		&\textup{THEN}~y_r^1(\bm{x}) = b_{r,0}^1+\sum_{d=1}^{D}b_{r,d}^1\cdot x_d\\
&\hspace*{15mm}\vdots\\
&\hspace*{10mm}y_r^C(\bm{x}) = b_{r,0}^C+\sum_{d=1}^{D}b_{r,d}^C\cdot x_d
	\end{split} \label{eq:rule}
\end{align}
where $\tilde{X}_{r, d}$ ($r=1,...,R$; $d=1,...,D$) is a Gaussian membership function (MF) for the $d$-th antecedent in the $r$-th rule, and $b_{r,d}^c$ ($c=1,...,C$) are the consequent parameters for the $c$-th class.

Let $V_{r,d}$ be the center of the Gaussian MF $\tilde{X}_{r,d}$, and $\Sigma_{r,d}$ be the corresponding standard deviation. Then, the membership grade of $x_{n,d}$ on $\tilde{X}_{r,d}$ is:
\begin{align}\label{eq:ms}
  \mu_{\tilde{X}_{r,d}}(x_{n,d})=\exp\left(\frac{-(x_{n,d}-V_{r,d})^2}{2\Sigma_{r,d}}\right),
\end{align}
and its firing level on the $r$-th rule is
\begin{align}
f_{n,r}=\prod_{d=1}^D \mu_{\tilde{X}_{r,d}}(x_{n,d}).
\end{align}

The normalized firing level is:
\begin{align}\label{eq:fl}
 \bar{f}_{n,r} = f_{n,r}\left/\sum_{i=1}^R f_{n,i}\right.,
\end{align}
and the estimate for the $c$-th class, $\hat{y}^c(\bm{x}_n)$, is
\begin{align}
\hat{y}^c(\bm{x}_n)=\sum_{r=1}^R y_r^c(\bm{x}_n)\bar{f}_{n,r}.
\end{align}

Once the antecedent parameters are initialized, LSE can be used to estimate the consequent parameters.

For a zero-order TSK fuzzy classifier, i.e., $b_{r,0}^c$ is adjustable whereas $b_{r,d}^c=0$ ($r=1,...,R; d=1,...,D; c=1,...,C$), we can define
\begin{align}
\hat{\bm{x}}_n&=[\bar{f}_{n,1}, \bar{f}_{n,2},...,\bar{f}_{n,R}]\in \mathbb{R}^{1\times  R}\\
\hat{X}&=[\hat{\bm{x}}_1;\cdots;\hat{\bm{x}}_N]\in \mathbb{R}^{N\times  R}\label{eq:hatX}\\
B&=[b_{1,0}^1,\ldots,b_{1,0}^C;\cdots;b_{R,0}^1,\ldots,b_{R,0}^C] \in \mathbb{R}^{R\times  C}
\end{align}
Then, the one-hot coding matrix $Y$ is estimated by
\begin{equation}\label{eq:pred}
	\hat{Y}=\hat{X}B,
\end{equation}
and the optimal $B$ can be solved by LSE, i.e.,
\begin{equation}
B=(\hat{X}^T\hat{X}+\lambda I)^{-1}\hat{X}^TY, \label{eq:B}
\end{equation}
where $\lambda$ is the weight of $L_2$ regularization in LSE.

For first-order TSK fuzzy classifier, we can define
\begin{align}
\hat{\bm{x}}_n&=[\bar{f}_{n,1}, \bar{f}_{n,1}\bm{x}_n,\bar{f}_{n,2},\bar{f}_{n,2}\bm{x}_n,\nonumber\\
&\qquad \cdots,\bar{f}_{n,R},\bar{f}_{n,R}\bm{x}_n]\in \mathbb{R}^{1\times  R(D+1)}\\
\hat{X}&=[\hat{\bm{x}}_1;\cdots;\hat{\bm{x}}_N]\in \mathbb{R}^{N\times  R(D+1)}\\
B_r&=\left[\begin{array}{cccc}
          b_{1,0}^1, & b_{1,0}^2, & \cdots, & b_{1,0}^C \\
          b_{1,1}^1, & b_{1,1}^2, & \cdots, & b_{1,1}^C \\
          \vdots & \vdots & \ddots & \vdots \\
          b_{1,D}^1, & b_{1,D}^2, & \cdots, & b_{1,D}^C
        \end{array}\right] \in \mathbb{R}^{(D+1)\times  C}\\
B&=[B_1;\cdots; B_R] \in \mathbb{R}^{R(D+1)\times  C}
\end{align}
Then, the optimal $B$ is still solved by  (\ref{eq:B}).

\subsection{FCM}

FCM~\cite{bezdek1984fcm} can be used to initialize the antecedent part of the $R$ rules of a multi-class TSK fuzzy classifier:
\begin{equation}
\begin{aligned}
  \min_{U,V}&~\sum_{n=1}^N \sum_{r=1}^R U_{n,r}^m \|\bm{x}_n-\bm{v}_r\|^2_2\\
  s.t.&~U_{n,r}\in [0,1]\\
   &~\sum_{r=1}^R U_{n,r}=1,\, n=1,...,N
\end{aligned},\label{eq:fcm}
\end{equation}
where $U_{n,r}$ is the membership grade of $\bm{x}_n$ in the $r$-th cluster, $\bm{v}_r$ (the $r$-th row of $V$) the center of the $r$-th cluster, and $m$ the fuzzy index.

Once $U$ and $V$ are obtained, the standard deviation of $\tilde{X}_{r,d}$ can be estimated by:
\begin{align}\label{eq:fs}
  &\Sigma_{r,d}=h\left[\sum_{n=1}^N U_{n,r} (x_{n,d}-V_{r,d})^2\left/\sum_{n=1}^N U_{n,r}\right.\right]^{1/2},
\end{align}
where $h$ is an adjustable scaling parameter.

\subsection{Integration of FCM and SSC}

When the data dimensionality is high, the pairwise distances $\|\bm{x}_n-\bm{v}_r\|^2_2$ in (\ref{eq:fcm}) become similar for different $n$ and $r$. As a result, the cluster centers obtained by the original FCM converge to the center of the input data space~\cite{winkler2011fuzzy}, losing their discriminability. To remedy this problem, SSC uses weighted Euclidean distances, i.e.,
\begin{equation}
	\|\bm{x}_n-\bm{v}_r\|_{\bm{w}_r}^2= \sum_{d=1}^D w_{r,d}(x_{n,d}-v_{r,d})^2,
\end{equation}
where $\bm{w}_r\in \mathbb{R}^{1\times D}$ is a weight vector associated with $\bm{v}_r$.

The cost function of AWFCM~\cite{keller2000fuzzy}, using weighted Euclidean distances, is:
\begin{equation}
	J_1(V, W, U)  = \sum_{n=1}^N\sum_{r=1}^RU_{n,r}^m\|\bm{x}_n-\bm{v}_r\|_{\bm{w}_r}^2,
\end{equation}
where $W=[\bm{w}_1;\ldots;\bm{w}_R]\in\mathbb{R}^{R\times D}$.

The cost function of SCAD~\cite{frigui2004unsupervised}, which further adds an exponential term $t$ to the weights, is:
\begin{equation}
	J_2(V, W, U)  = \sum_{n=1}^N\sum_{r=1}^RU_{n,r}^m\|\bm{x}_n-\bm{v}_r\|_{\bm{w}_r^t}^2.
\end{equation}

Zhou \emph{et al.}~\cite{zhou2016fuzzy} proposed EWFCM, which uses entropy as a regularization term to control the weights, and modifies the cost function of FCM to:
\begin{equation}
\begin{aligned}
	J_3(V, W, U)=&\sum_{n=1}^N\sum_{r=1}^R U_{n,r}^m\|\bm{x}_n-\bm{v}_r\|_{\bm{w}_r}^2 \\&+\gamma \sum_{r=1}^R\sum_{d=1}^D W_{r,d}\ln(W_{r,d})
\end{aligned}
\end{equation}
where $\gamma$ is a user-specified regularization parameter.

In addition to the within-cluster compactness, the between-cluster separation is also very important in clustering~\cite{wu2005novel}. ESSC~\cite{deng2010enhanced} takes the between-cluster separation into consideration, and its cost function is
\begin{equation}\label{eq:essc}
\begin{aligned}
	J_4(V, W, U)=&\sum_{n=1}^N\sum_{r=1}^RU_{n,r}^m\|\bm{x}_n-\bm{v}_r\|_{\bm{w}_r}^2 \\
	&+\gamma \sum_{r=1}^R\sum_{d=1}^D W_{r,d}\ln(W_{r,d})
	\\ &- \eta \sum_{r=1}^R \left(\sum_{n=1}^NU_{n,r}^m\right)\|\bm{v}_r-\bm{v}_0\|_{\bm{w}_r}^2,
\end{aligned}
\end{equation}
where $\bm{v}_0=\frac{1}{N}\sum_{n=1}^N\bm{x}_n$ is the center of all data samples, and $\eta \in [0,1)$ is a user-specified regularization parameter.

\subsection{Supervised Fuzzy Partition (SFP)}

Although SSC can be used to learn a better cluster structure than the traditional FCM, it does not utilize the label information in supervised learning. Ideally, a cluster should only contain data samples with the same  label. Recently, supervised fuzzy partition (SFP)~\cite{ashtari2020supervised} was proposed to tackle this problem. Its objective function integrates SSC and label information:
\begin{equation}
\begin{aligned}
	\min_{U,V,W,Z} &\sum_{n=1}^N\sum_{r=1}^RU_{n,r}\|\bm{x}_n-\bm{v}_r\|_{w_r}^2\\
	&+\alpha \sum_{n=1}^N\sum_{r=1}^RU_{n,r}\ell(\bm{y}_n,\bm{z}_r)\\
	&+\gamma \sum_{n=1}^N\sum_{r=1}^RU_{n,r}\ln(U_{n,r})\\
	&+\lambda \sum_{r=1}^R\sum_{d=1}^DW_{r,d}\ln(W_{r,d})\\
	s.t.&~U_{n,r}\in [0,1], \quad 	~W_{r,d}\in [0,1]\\
	&~\sum_{r=1}^RU_{n,r}=1, \quad 	~\sum_{d=1}^DW_{r,d}=1,\\
\end{aligned}
\end{equation}
where $\bm{z}_r\in \mathbb{R}^{1\times C}$ indicates the label prediction for Cluster $r$, and $\ell$ a loss function.

In addition to the utilization of label information, SFP uses the entropy, instead of the fuzzy index $m$, to regularize the membership grades $U$. However, it does not consider the between-cluster separation, compared with ESSC.


\subsection{SESSC}

We propose SESSC to integrate the label information with ESSC, which uses the following cost function to constrain the labels in each cluster, following~\cite{ashtari2020supervised}:
\begin{equation}\label{eq:logloss}
	J_l=\sum_{n=1}^N\sum_{r=1}^RU_{n,r}^m\left[-\sum_{c=1}^C Y_{n,c}\ln(Z_{r,c})\right],
\end{equation}
where $Z_{r,c}$ indicates the probability that Cluster~$r$ belongs to Class~$c$. Clearly, $\sum_{c=1}^C Z_{r,c}=1$, $r=1,...,R$.

An equivalence of (\ref{eq:logloss}) is:	
\begin{equation}\label{eq:mseloss}
	J_l'=\sum_{n=1}^N\sum_{r=1}^R U_{n,r}^m\|\bm{y}_n-\bm{z}_r\|_2^2,
\end{equation}
where $\bm{z}_r=[Z_{r,1},...,Z_{r,C}]$. (\ref{eq:mseloss}) can be easily extended to regression problems.

Minimizing (\ref{eq:logloss}) or (\ref{eq:mseloss}) leads to the optimal label $Z_{r,c}$:
\begin{align}
	Z_{r,c}=\frac{Z'_{r,c}}{\sum_{i=1}^R Z'_{i,c}}, \label{sol:zc}
\end{align}
where
\begin{align}
Z'_{r,c}=\frac{\sum_{n=1}^NU_{n,r}^m\cdot \bm{y}_{n,c}}{\sum_{n=1}^NU_{n,r}^m}.
\end{align}

Adding (\ref{eq:logloss}) to the cost function of ESSC, we obtain the cost function for our proposed SESSC:
\begin{equation}\label{eq:sessc}
	\begin{aligned}
		\min_{V,W,U,Z} &\sum_{n=1}^N\sum_{r=1}^RU_{n,r}^m\|\bm{x}_n-\bm{v}_r\|_{\bm{w}_r}^2 \\
	&+\gamma \sum_{r=1}^R\sum_{d=1}^D W_{r,d}\ln(W_{r,d}) \\
	&- \eta \sum_{r=1}^R \left(\sum_{n=1}^NU_{n,r}^m\right)\|\bm{v}_r-\bm{v}_0\|_{\bm{w}_r}^2\\
	&+\beta \sum_{n=1}^N\sum_{r=1}^RU_{n,r}^m\left[-\sum_{c=1}^C Y_{n,c}\ln(Z_{r,c})\right]\\
	s.t.&~U_{n,r}\in [0,1], \quad 	~W_{r,d}\in [0,1]\\
	&~\sum_{r=1}^RU_{n,r}=1, \quad 	~\sum_{d=1}^DW_{r,d}=1.\\
	\end{aligned}
\end{equation}

\subsection{Solution of SESSC}

(\ref{eq:sessc}) can be solved using the Lagrange multiplier method. We first form the following Lagrangian function:
\begin{equation}
	\begin{aligned}
		J(W,V,U,Z,\bm{\alpha},\bm{\zeta})=&\sum_{n=1}^N\sum_{r=1}^RU_{n,r}^m\|\bm{x}_n-\bm{v}_r\|_{\bm{w}_r}^2 \\
	&+\gamma \sum_{r=1}^R\sum_{d=1}^D W_{r,d}\ln(W_{r,d}) \\
	&- \eta \sum_{r=1}^R \left(\sum_{n=1}^NU_{n,r}^m\right)\|\bm{v}_r-\bm{v}_0\|_{\bm{w}_r}^2\\
	&+\beta \sum_{n=1}^N\sum_{r=1}^RU_{n,r}^m\left[-\sum_{c=1}^CY_{r,c}\ln(Z_{r,c})\right] \\
	&+ \sum_{n=1}^N \alpha_n \left(\sum_{r=1}^RU_{n,r}-1\right)\\
	& + \sum_{r=1}^R\zeta_r \left(\sum_{d=1}^DW_{r,d}-1\right)
	\end{aligned}
\end{equation}
where $\bm{\alpha}=[\alpha_1,...,\alpha_N]$ and $\bm{\zeta}=[\zeta_1,...,\zeta_R]$ are Lagrangian multiplier coefficients.

When $V$, $W$ and $Z$ are fixed, we can compute the optimal value of $U$ by setting $\partial J/\partial U_{n,r} = 0$  and  $\partial J/\partial \alpha_n = 0$, i.e.,
\begin{align}
	\frac{\partial J}{\partial U_{n,r}}=&mU_{n,r}^{m-1}\{\|\bm{x}_n-\bm{v}_r\|_{\bm{w}_r}^2 - \eta \|\bm{v}_r-\bm{v}_0\|_{\bm{w}_r}^2 \nonumber \\
	&+ \beta [-\sum_{c=1}^CY_{r,c}\ln(Z_{r,c})]\} - \alpha_n = 0 \label{sol:u1}\\
	\frac{\partial J}{\partial \alpha_n}=&\sum_{r=1}^RU_{n,r}-1=0,\label{sol:u2}
\end{align}
which lead to
\begin{equation}
	U_{n,r}=\frac{D_{n,r}^{-1/(m-1)}}{\sum_{i=1}^RD_{n,i}^{-1/(m-1)}},\label{sol:u}
\end{equation}
where
\begin{align}
	D_{n,r}=&\max\{0, \|\bm{x}_n-\bm{v}_r\|_{\bm{w}_r}^2 - \eta \|\bm{v}_r-\bm{v}_0\|_{\bm{w}_r}^2 \nonumber \\
&- \beta \sum_{c=1}^CY_{r,c}\ln(Z_{r,c})\}.
\end{align}

When $U$, $W$ and $Z$ are fixed, we can compute the optimal value of $V$ by setting $\partial J/\partial V_{r,d} = 0$, i.e.,
\begin{equation}\label{sol:v1}
	\frac{\partial J}{\partial V_{r,d}}=-W_{r,d}\sum_{n=1}^NU_{n,r}^m\left[X_{n,d}-\eta V_{0,d} - (1-\eta) V_{r,d}\right]=0,
\end{equation}
which leads to
\begin{equation}\label{sol:v}
	V_{r,d}=\frac{\sum_{n=1}^NU_{n,r}(X_{n,d}-\eta V_{0,d})}{\sum_{n=1}^NU_{n,r}(1-\eta)}.
\end{equation}

When $U$, $V$ and $Z$ are fixed, we can compute the optimal value of $W$ by setting $\partial J / \partial W_{r,d} = 0$ and $\partial J/\partial \zeta_r=0$, i.e.,
\begin{align}
	\frac{\partial J}{\partial W_{r,d}}=&\sum_{n=1}^NU_{n,r}^m\left[(X_{n,d}-V_{r,d})^2-\eta (V_{r,d}-V_{0,d})^2\right]\nonumber\\
	&+ \gamma \ln(W_{r,d})+\gamma -\zeta_r \label{sol:w1}\\
	\frac{\partial J}{\partial \zeta_r}=&\sum_{d=1}^DW_{r,d}-1=0\label{sol:w2}
\end{align}
which lead to
\begin{equation}\label{sol:w}
	W_{r,d}=\frac{\exp(-S_{r,d}/\gamma)}{\sum_{d'=1}^D\exp(-S_{r,d'}/\gamma)},
\end{equation}
where
\begin{align}
	S_{r,d}=\sum_{n=1}^NU_{n,r}^m\left[(X_{n,d}-V_{r,d})^2-\eta (V_{r,d}-V_{0,d})^2\right].
\end{align}

When $U$, $V$ and $W$ are fixed, the label of the $r$-th cluster, $\bm{z}_r$, can be computed by (\ref{sol:zc}).

The pseudo-code of SESSC is summarized in Algorithm~\ref{alg:sessc}. The center matrix $V$ is initialized from $k$-means clustering.

\begin{algorithm}
\KwIn{Input data matrix, $X\in\mathbb{R}^{N\times D}$\;
 \hspace*{10mm} Input label matrix, $Y\in\mathbb{R}^{N\times C}$\;
 \hspace*{10mm} Number of clusters, $R$\;
 \hspace*{10mm} Regularization weights $\gamma$, $\eta$ and $\beta$\;
 \hspace*{10mm} Maximum number of iterations, $T$\;
 \hspace*{10mm} Error threshold to stop iteration, $\epsilon$.
}
\KwOut{Cluster center matrix $V$\;
 \hspace*{12mm} Cluster weight matrix $W$\;
 \hspace*{12mm} Membership grade matrix $U$\;
 \hspace*{12mm} Cluster label matrix $Z$.}	
 Initialize $V$ by $k$-means ($k=R$) clustering on $X$\;
 $\bm{v}_0=\frac{1}{N}\sum_{n=1}^N\bm{x}_n$\;
 $\hat{V}= V$\;
 Initialize $W_{r,d} = 1/D,\ r=1,...,R;\ d=1,...,D$\;
 Initialize $Z_{r,c} = 1/C,\ r=1,...,R;\ c=1,...,C$\;
   \For{$t=1:T$}{
   Update $U$ by (\ref{sol:u})\;
   Update $V$ by (\ref{sol:v})\;
   Update $W$ by (\ref{sol:w})\;
   Update $Z$ by (\ref{sol:zc})\;
   \eIf{$\|\hat{V}-V\|_2 < \epsilon$}{
   	break\;
   }{
   $\hat{V}= V$\;
   }
   }
\caption{Pseudo-code of the SESSC.}\label{alg:sessc}
\end{algorithm}

\begin{figure}[htpb]\centering
\includegraphics[width=0.9\columnwidth,clip]{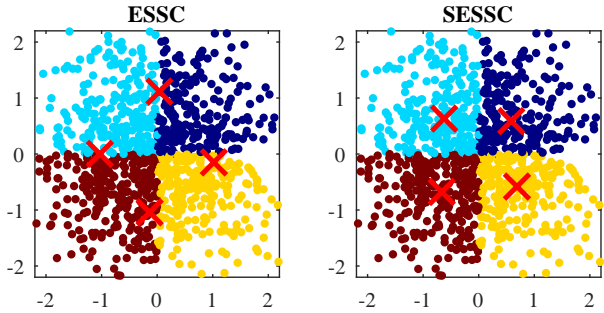}
	\caption{ESSC and SESSC clustering results on a four-class classification problem with $R=4$. Dots in different colors represent data samples with different labels. Cluster centers are indicated by red $\bm{\times}$.} \label{fig:xor}
\end{figure}

A toy example illustrating a four-class classification problem and the corresponding cluster centers obtained from ESSC and SESSC are shown in Fig.~\ref{fig:xor}. The data samples were generated from a 2-dimensional Gaussian distribution $\bm{N}(0, 1)$, and each quadrant represents a different class. Clearly, the cluster centers generated from SESSC are more reasonable, and hence may lead to better classification performance when used to initialize a TSK fuzzy classifier.

\subsection{SESSC as a Classifier}\label{subsec:sessctsk}

SESSC itself can be used as a classifier. Following the approach used in \cite{ashtari2020supervised}, for a test sample $\bm{x}_t$, we can estimate the memberships by:
\begin{equation}\label{sol:utest}
	U'_{t,r}=\frac{D_{t,r}'^{-1/(m-1)}}{\sum_{i=1}^RD_{t,i}'^{-1/(m-1)}},
\end{equation}
where
\begin{equation}
	D_{t,r}'=\|\bm{x}_t-\bm{v}_r\|_{\bm{w}_r}^2 - \eta \|\bm{v}_r-\bm{v}_0\|_{\bm{w}_r}^2.
\end{equation}
Then, the label matrix $\hat{Y}$ is computed as
\begin{align}\label{sol:label}
    \hat{Y} = \hat{Y}'\left/\sum_{c=1}^C \hat{Y}'_{\cdot,c}\right.,
\end{align}
where
\begin{align}
	\hat{Y}'=U'Z.
\end{align}

\subsection{SESSC-LSE}

SESSC can also be used to initialize the antecedent part of a TSK fuzzy classifier, and then LSE can be used to estimate the consequent parameters. This approach is denoted SESSC-LSE in this paper. Its pseudo-code is given in Algorithm~\ref{alg:sessctsk}.

\begin{algorithm}
\KwIn{Training data $X\in\mathbb{R}^{N\times D}$\;
 \hspace*{10mm} Training label matrix $Y\in\mathbb{R}^{N\times C}$\;
 \hspace*{10mm} Test data $X_t\in\mathbb{R}^{N_t\times D}$\;
 \hspace*{10mm} Number of clusters, $R$\;
 \hspace*{10mm} Regularization weight parameters $\gamma$, $\eta$ and $\beta$\;
 \hspace*{10mm} Maximum number of iterations, $T$\;
 \hspace*{10mm} Error threshold to stop iteration, $\epsilon$\;
 \hspace*{10mm} Weight of $L_2$ regularization, $\lambda$\;
 \hspace*{10mm} Scaling parameter $h$\;
}
\KwOut{Predicted label matrix $\hat{Y}_t$\;}	

Compute $V$ and $U$ by SESSC in Algorithm~\ref{alg:sessc}\;
Compute $\Sigma$ by (\ref{eq:fs})\;
Compute $\bar{f}$ and $\bar{f}_t$ for the training data and test data, respectively, by (\ref{eq:fl})\;
Compute $\hat{X}$ by (\ref{eq:hatX}) using $\bar{f}$\;
Estimate the consequent parameters $B$ by (\ref{eq:B})\;
Compute $\hat{X}_t$ by (\ref{eq:hatX}) using $\bar{f}_t$\;
Compute the prediction matrix $\hat{Y}_t$ by (\ref{eq:pred}) using $\hat{X}_t$\;
\caption{The SESSC-LSE algorithm.}\label{alg:sessctsk}
\end{algorithm}

\section{Experiments}\label{sec:exp}

This section evaluates the performance of our proposed SESSC and SESSC-LSE on both synthetic and real-world datasets. Both zero-order and first-order TSK fuzzy classifiers were considered.

\subsection{Synthetic 2-Dimensional Datasets}

\begin{figure}[htpb]
\hspace*{-.15\linewidth}\subfigure[]{\includegraphics[width=1.25\linewidth,clip]{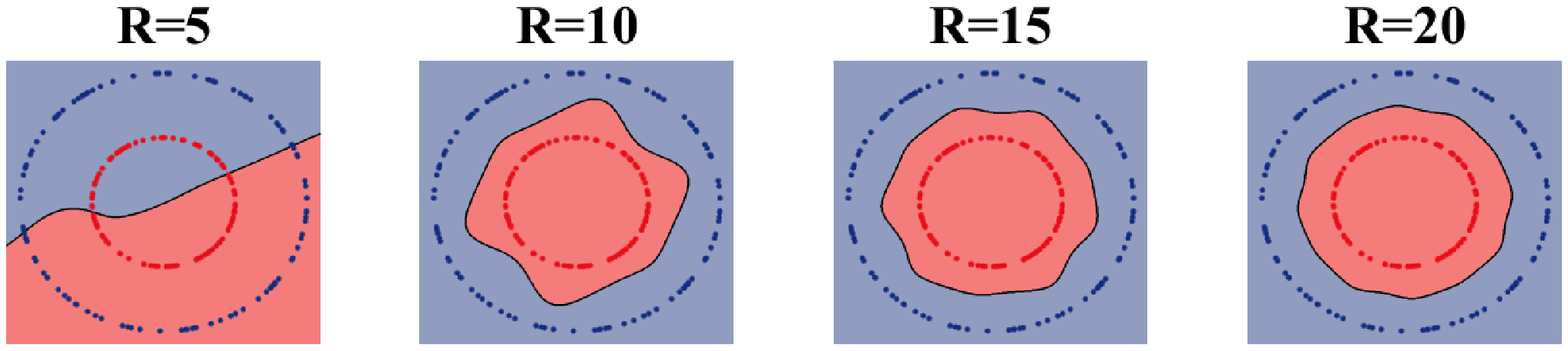}}
\hspace*{-.15\linewidth}\subfigure[]{\includegraphics[width=1.25\linewidth,clip]{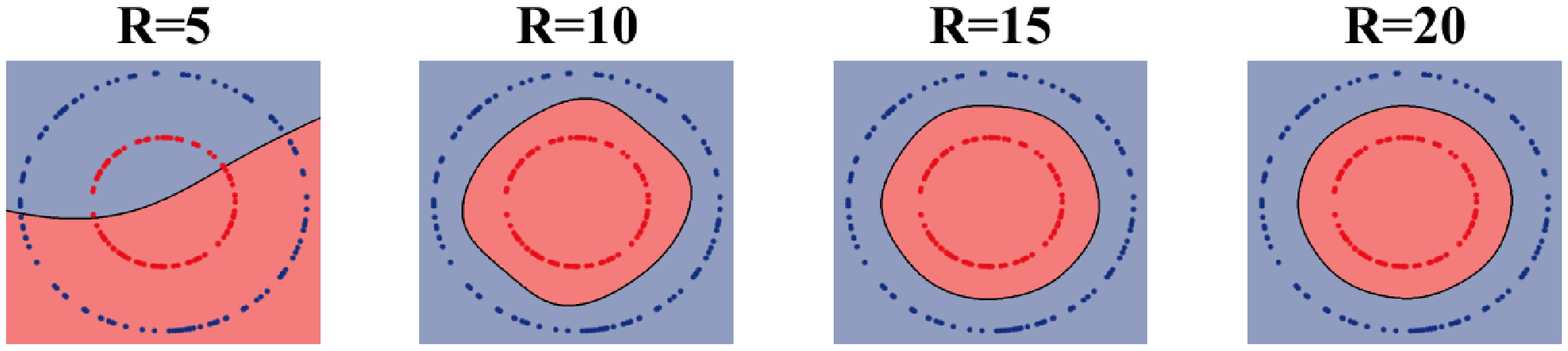}}
\hspace*{-.15\linewidth}\subfigure[]{\includegraphics[width=1.25\linewidth,clip]{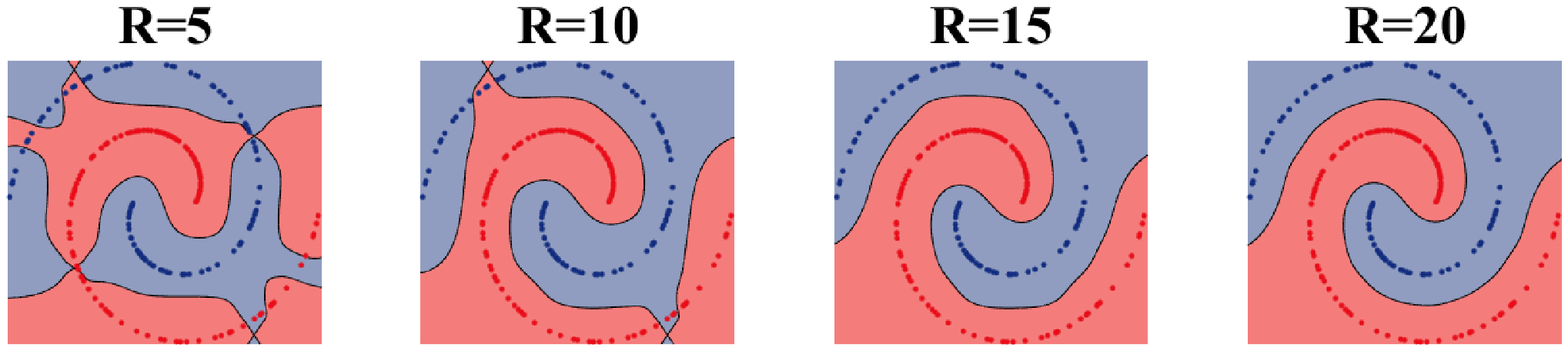}}
\hspace*{-.15\linewidth}\subfigure[]{\includegraphics[width=1.25\linewidth,clip]{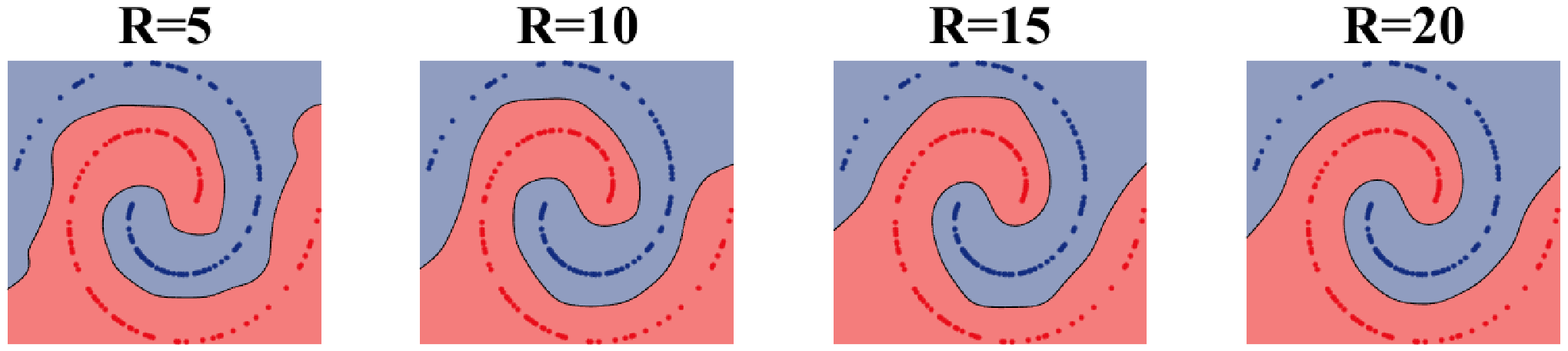}}
	\caption{Decision boundaries on two synthetic 2-dimensional datasets. (a) SESSC on concentric circles; (b) SESSC-LSE  on concentric circles; (c) SESSC on spiral; (d) SESSC-LSE  on spiral.}\label{fig:2ddata}
\end{figure}

First, we evaluated the performances of our proposed SESSC and SESSC-LSE on two synthetic 2-dimensional datasets, concentric circles and spiral. The parameters of SESSC were $R=\{5,10,15,20\}$, $\gamma=100$, $\eta=0.01$ and $\beta=0.1$. Zero-order SESSC-LSE was used here for a fair comparison. Additional parameters for SESSC-LSE were $h=1$ and $\lambda=0.01$. The results are shown in Fig.~\ref{fig:2ddata}, in which the dots represent the generated data samples, and the black curves the decision boundaries learned by SESSC and SESSC-LSE.

Fig.~\ref{fig:2ddata} shows that:
\begin{enumerate}
	\item When $R$, the number of clusters, increased, the SESSC decision boundaries became smoother and closer to the true decision boundaries. This demonstrated that our proposed SESSC can fit smooth decision boundaries and achieve good generalization performance.
	\item For the same $R$, SESSC-LSE outperformed SESSC, which indicated that SESSC-LSE can further improve the classification performance.
\end{enumerate}

\subsection{Real-World High-Dimensional Datasets}

Next, we evaluated our proposed algorithms on nine classification datasets from the UCI Machine Learning Respository\footnote{http://archive.ics.uci.edu/ml/index.php}. Their characteristics are summarized in Table~\ref{tab:data}. For each dataset, we randomly selected 70\% samples as the training set and the remaining 30\% as the test set 30 times to get 30 different data splits. We ran each algorithm on these 30 data splits and report the average performance.

\begin{table}[htpb]\centering
\begin{threeparttable} \setlength{\tabcolsep}{1.2mm}{
\caption{Summary of the nine UCI datasets.} \label{tab:data}
\begin{tabular}{cccc}\toprule
 Dataset    & $N$, no. of samples & $D$, no. of features & $C$, no. of classes \\\midrule
 WPBC\tnote{1}       & 198            & 32              & 2              \\
 WDBC\tnote{2}       & 569            & 30              & 2              \\
 Vehicle\tnote{3}    & 846            & 18              & 4              \\
 Biodeg\tnote{4}     & 1,055           & 41              & 2              \\
 DRD\tnote{5}        & 1,151           & 19              & 2              \\
 Steel\tnote{6}      & 1,941           & 27              & 7              \\
 IS\tnote{7}         & 2,310           & 19              & 7              \\
 Waveform21\tnote{8} & 5,000           & 21              & 3              \\
 Satellite\tnote{9}  & 6,435           & 36              & 6             \\\bottomrule
\end{tabular}}
\begin{tablenotes}
\item[1] https://archive.ics.uci.edu/ml/datasets/Breast+Cancer+Wisconsin+ (Prognostic)
\item[2] https://archive.ics.uci.edu/ml/datasets/Breast+Cancer+Wisconsin+ (Diagnostic)
\item[3] https://archive.ics.uci.edu/ml/datasets/Statlog+\%28Vehicle+ Silhouettes\%29
\item[4] https://archive.ics.uci.edu/ml/datasets/QSAR+biodegradation
\item[5] https://archive.ics.uci.edu/ml/datasets/Diabetic+Retinopathy+Debrecen +Data+Set
\item[6] https://archive.ics.uci.edu/ml/datasets/Steel+Plates+Faults
\item[7] https://archive.ics.uci.edu/ml/datasets/Image+Segmentation
\item[8] https://archive.ics.uci.edu/ml/datasets/Waveform+Database+Generator +(Version+1)
\item[9] https://archive.ics.uci.edu/ml/datasets/Statlog+(Landsat+Satellite)
\end{tablenotes}
\end{threeparttable}
\end{table}

Some datasets contain both numerical and categorical features. The categorical features were converted to numerical ones by one-hot encoding. Then, all features were normalized by $z$-score using mean and standard deviation computed from the training set.

\subsection{Performance Measures}

Since some datasets have significant class imbalance, both raw classification accuracy (RCA) and balanced classification accuracy (BCA) were used as the performance measure:
\begin{itemize}
	\item RCA: The total number of correct classifications divided by the total number of samples.
	\item BCA: The average of per-class RCAs, which is less affected by class imbalance.
\end{itemize}

\subsection{Algorithms}

We used LSE to compute the consequent parameters for all TSK fuzzy classifiers [see (\ref{eq:B})]. We replaced SESSC in SESSC-LSE by three different clustering algorithms (FCM, EWFCM, and ESSC) to compare their performances with SFP and the proposed SESSC and SESSC-LSE. Because SFP and SESSC are zero-order fuzzy classifiers, we did not compare them with first-order fuzzy classifier (FCM-LSE, EWFCM-LSE, ESSC-LSE and SESSC-LSE).

The values or search ranges of the parameters in different algorithms are shown in Table~\ref{tab:para}. We set $R=30$ to keep the fuzzy classifiers concise so they can have interpretability. Note that in the original SFP paper \cite{ashtari2020supervised}, the number of clusters $R$ was optimized in $[D, N]$, which was much larger than 30 used in our experiments. This might be the reason why SFP performed worse than other SSC based fuzzy classifiers. In the actual implementation~\cite{ashtari2020supervised}, SFP used modified parameters $\gamma'$, $\alpha'$, $\lambda'$ during grid research, where $\gamma=(1-\gamma')/\gamma'$, $\alpha=(1-\alpha')/\alpha'$ and $\lambda=(1-\lambda')/\lambda'$.

\begin{table*}[h]\centering \setlength{\tabcolsep}{5mm}
\caption{Parameter settings in the algorithms.}\label{tab:para}
\begin{tabular}{c|cc}\toprule
Parameter     &  Value or search range         & Used in \\\midrule
$R$ & 30 & All algorithms   \\
$m$ & $\frac{\min(N, D-1)}{\min(N, D-1)-2}$ if $\min(N, D-1)>2$, otherwise 2  & All algorithms except SFP     \\
$h$ &  $\{0.01, 0.1, 1, 10, 100\}$ & All algorithms except SFP and SESSC   \\
$\lambda$ & $\{0.0001, 0.001, 0.01, 0.1, 1, 10, 100\}$ & All algorithms except SFP and SESSC   \\
$\gamma$ & $\{0.01, 0.1, 1, 10, 100\}$ &  EWFCM-LSE, ESSC-LSE, SESSC, SESSC-LSE  \\
$\eta$ & $\{0.01, 0.05, 0.1, 0.3, 0.5\}$ & ESSC-LSE, SESSC, SESSC-LSE\\
$\beta$ & $\{0.01, 0.1, 1, 10, 100\}$ & SESSC, SESSC-LSE\\
$\gamma'$ & $\{0.55, 0.65, 0.75, 0.85, 0.95\}$ & SFP \\
$\alpha'$ & $\gamma'/2$ & SFP \\
$\lambda'$ & $\{0.15, 0.25, 0.35, 0.45, 0.55, 0.65,0.75, 0.85, 0.95\}$& SFP \\ \bottomrule
\end{tabular}
\end{table*}

We tuned all parameters using 5-fold cross validation on the training set and chose the combination with the highest average BCA. The best $\gamma$, $\eta$ and $\beta$ from SESSC were used in SESSC-LSE.

\subsection{Experimental Results}\label{subsec:expres}

The average test RCAs and BCAs of zero-order fuzzy classifiers are shown in Table~\ref{tab:expres}. The best performance on each dataset is marked in bold. The ranks of the average RCAs and BCAs are shown in Table~\ref{tab:exprank}.

\begin{table*}[!h]\centering  \setlength{\tabcolsep}{0.8mm}
\caption{Average RCAs and BCAs of the five zero-order algorithms on the nine datasets.}\label{tab:expres}
\begin{tabular}{c|cccccc|cccccc} \toprule
           & \multicolumn{6}{c|}{RCA}                                                      & \multicolumn{6}{c}{BCA}                                                \\\midrule
Dataset    & FCM-LSE        & EWFCM-LSE & ESSC-LSE       & SFP   & SESSC & SESSC-LSE      & FCM-LSE        & EWFCM-LSE & ESSC-LSE & SFP   & SESSC & SESSC-LSE      \\\midrule
WPBC       & \textbf{77.00} & 74.11     & 75.39          & 75.89 & 64.06 & 69.72          & 51.54          & 50.57     & 54.22    & 56.12 & 56.19 & \textbf{56.32} \\
WDBC       & 94.05          & 93.90     & \textbf{95.75} & 93.94 & 95.44 & 95.38          & 92.85          & 93.00     & 95.08    & 92.82 & 95.09 & \textbf{95.30} \\
Vehicle    & 70.85          & 70.17     & 70.28          & 68.62 & 67.49 & \textbf{71.93} & 71.21          & 70.53     & 70.63    & 68.91 & 67.75 & \textbf{72.25} \\
Biodeg     & 66.47          & 80.78     & 81.05          & 78.17 & 82.60 & \textbf{85.64} & 50.95          & 78.31     & 78.07    & 70.45 & 81.56 & \textbf{82.91} \\
DRD        & 64.47          & 64.40     & 64.85          & 61.02 & 61.50 & \textbf{66.06} & 64.61          & 64.66     & 65.23    & 60.02 & 61.35 & \textbf{66.34} \\
Steel      & 64.09          & 65.32     & 66.12          & 55.50 & 67.26 & \textbf{70.68} & 56.82          & 59.56     & 60.17    & 32.11 & 67.70 & \textbf{69.47} \\
IS         & 86.83          & 83.94     & 84.16          & 83.98 & 86.68 & \textbf{88.92} & 86.83          & 83.94     & 84.16    & 83.98 & 86.68 & \textbf{88.92} \\
Waveform21 & \textbf{86.55} & 85.66     & 86.07          & 76.89 & 83.51 & 86.50          & \textbf{86.52} & 85.63     & 86.03    & 76.82 & 83.47 & 86.47          \\
Satellite  & 44.62          & 86.47     & 86.19          & 84.80 & 86.21 & \textbf{86.94} & 39.56          & 82.66     & 82.05    & 79.95 & 83.28 & \textbf{84.27} \\\midrule
Average    & 72.77          & 78.31     & 78.87          & 75.42 & 77.19 & \textbf{80.20} & 66.77          & 74.32     & 75.07    & 69.02 & 75.90 & \textbf{78.03}
\\ \bottomrule
\end{tabular}
\end{table*}

\begin{table*}[!h]\centering \setlength{\tabcolsep}{0.8mm}
\caption{RCA and BCA ranks of the five zero-order algorithms on the nine datasets.}\label{tab:exprank}
\begin{tabular}{c|cccccc|cccccc}\toprule
                      & \multicolumn{6}{c|}{RCA rank}                                     & \multicolumn{6}{c}{BCA rank}                                     \\\midrule
Dataset    & FCM-LSE & EWFCM-LSE & ESSC-LSE & SFP & SESSC & SESSC-LSE    & FCM-LSE & EWFCM-LSE & ESSC-LSE & SFP & SESSC & SESSC-LSE    \\\midrule
WPBC       & 1       & 4         & 3        & 2   & 6     & 5         & 5       & 6         & 4        & 3   & 2     & 1         \\
WDBC       & 4       & 6         & 1        & 5   & 2     & 3         & 5       & 4         & 3        & 6   & 2     & 1         \\
Vehicle    & 2       & 4         & 3        & 5   & 6     & 1         & 2       & 4         & 3        & 5   & 6     & 1         \\
Biodeg     & 6       & 4         & 3        & 5   & 2     & 1         & 6       & 3         & 4        & 5   & 2     & 1         \\
DRD        & 3       & 4         & 2        & 6   & 5     & 1         & 4       & 3         & 2        & 6   & 5     & 1         \\
Steel      & 5       & 4         & 3        & 6   & 2     & 1         & 5       & 4         & 3        & 6   & 2     & 1         \\
IS         & 2       & 6         & 4        & 5   & 3     & 1         & 2       & 6         & 4        & 5   & 3     & 1         \\
Waveform21 & 1       & 4         & 3        & 6   & 5     & 2         & 1       & 4         & 3        & 6   & 5     & 2         \\
Satellite  & 6       & 2         & 4        & 5   & 3     & 1         & 6       & 3         & 4        & 5   & 2     & 1         \\\midrule
Average    & 3.3     & 4.2       & 2.9      & 5.0 & 3.8   & \textbf{1.8}       & 4.0     & 4.1       & 3.3      & 5.2 & 3.2   & \textbf{1.1}
\\\bottomrule
\end{tabular}
\end{table*}

Tables~\ref{tab:expres} and\ref{tab:exprank} show that:
\begin{enumerate}
	\item For zero-order fuzzy classifiers, our proposed SESSC-LSE achieved the best RCA on six out of the nine datasets, and best BCA on eight out of the nine datasets. SESSC-LSE's performance was also close to the best on the remaining datasets. On average, SESSC-LSE achieved the best RCA and BCA, and also ranked the first.
	\item Our proposed SESSC and SESSC-LSE performed better on datasets with class imbalance. For instance, FCM-LSE and EWFCM-LSE achieved much higher RCAs on WPBC, but their BCAs were almost random (50\% BCA for binary classification). Although SESSC had a lower average RCA than EWFCM-LSE and ESSC-LSE, its average BCA was higher. The proposed SESSC-LSE further improved the BCA performance of SESSC.
	\item On average, all four SSC-based algorithms (EWFCM-LSE, ESSC-LSE, SFP and SESSC-LSE) outperformed FCM-LSE. This demonstrated that SSC-based clustering algorithms can result in clusters with higher discriminability on high-dimensional datasets.
\end{enumerate}

We also compared four first-order TSK classifiers and show their performances in Tables~\ref{tab:expreso1} and \ref{tab:expranko1}. All four of them performed similarly, and all three SSC-based classifiers had stable performances (first-order FCM-LSE performed significantly worse than SSC-based classifiers on the Satellite dataset). This suggests that when the consequent complexity of the TSK fuzzy classifier increases, the antecedent initialization may become less important.

\begin{table*}[!h]\centering
\caption{Average RCAs and BCAs of the four first-order algorithms on the nine datasets.}\label{tab:expreso1}
\begin{tabular}{c|cccc|cccc}\toprule
           & \multicolumn{4}{c|}{RCA}                                           & \multicolumn{4}{c}{BCA}                                           \\\midrule
Dataset    & FCM-LSE        & EWFCM-LSE      & ESSC-LSE       & SESSC-LSE      & FCM-LSE        & EWFCM-LSE      & ESSC-LSE       & SESSC-LSE      \\\midrule
WPBC      & \textbf{76.72} & 68.72          & 71.22          & 70.11          & \textbf{64.69} & 56.99          & 58.95          & 61.05          \\
WDBC       & 96.14          & 95.98          & \textbf{96.84} & 96.08          & 95.12          & 95.26          & \textbf{96.16} & 95.58          \\
Vehicle    & 83.31          & \textbf{83.67} & 83.32          & 82.49          & 83.50          & \textbf{83.85} & 83.50          & 82.69          \\
Biodeg     & 85.67          & 86.92          & 86.74          & \textbf{87.02} & 83.47          & \textbf{85.02} & 84.70          & 84.88          \\
DRD        & \textbf{71.10} & 71.06          & 70.36          & 70.01          & \textbf{71.38} & 71.36          & 70.69          & 70.29          \\
Steel      & 74.32          & 73.15          & 74.07          & \textbf{74.83} & 75.74          & 74.84          & 75.32          & \textbf{76.20} \\
IS         & \textbf{94.98} & 94.54          & 94.53          & 94.61          & \textbf{94.98} & 94.54          & 94.53          & 94.61          \\
Waveform21 & 86.80          & 86.75          & 86.83          & \textbf{86.90} & 86.76          & 86.72          & 86.80          & \textbf{86.87} \\
Satellite  & 84.33          & 90.17          & 90.40          & 90.03          & 77.24          & 87.51          & \textbf{87.76} & 87.63          \\\midrule
Average    & \textbf{83.71} & 83.44          & 83.81 & 83.56          & 81.43          & 81.79          & 82.05          & \textbf{82.20}\\\bottomrule
\end{tabular}	
\end{table*}

\begin{table*}[!h]\centering
\caption{RCA and BCA ranks of the four first-order algorithms on the nine datasets.}\label{tab:expranko1}
\begin{tabular}{c|cccc|cccc}\toprule
           & \multicolumn{4}{c|}{RCA rank}                         & \multicolumn{4}{c}{BCA rank}                       \\\midrule
Dataset    & FCM-LSE      & EWFCM-LSE & ESSC-LSE & SESSC-LSE & FCM-LSE & EWFCM-LSE & ESSC-LSE & SESSC-LSE    \\\midrule
WPBC       & 1            & 4         & 2        & 3         & 1       & 4         & 3        & 2            \\
WDBC       & 2            & 4         & 1        & 3         & 4       & 3         & 1        & 2            \\
Vehicle    & 3            & 1         & 2        & 4         & 2       & 1         & 2        & 4            \\
Biodeg     & 4            & 2         & 3        & 1         & 4       & 1         & 3        & 2            \\
DRD        & 1            & 2         & 3        & 4         & 1       & 2         & 3        & 4            \\
Steel      & 2            & 4         & 3        & 1         & 2       & 4         & 3        & 1            \\
IS         & 1            & 3         & 4        & 2         & 1       & 3         & 4        & 2            \\
Waveform21 & 3            & 4         & 2        & 1         & 3       & 4         & 2        & 1            \\
Satellite  & 4            & 2         & 1        & 3         & 4       & 3         & 1        & 2            \\\midrule
Average    & \textbf{2.3} & 2.9       & 2.3      & 2.4       & 2.4     & 2.8       & 2.4      & \textbf{2.2}\\\bottomrule
\end{tabular}
\end{table*}

\subsection{Statistical Analysis}

To further evaluate if the performance improvements of our proposed SESSC-LSE over others were statistically significant, we also performed non-parametric multiple comparison tests on the RCAs and BCAs using Dunn's procedure~\cite{dunn1964multiple}, with a $p$-value correction using the False Discovery Rate method~\cite{benjamini1995controlling}. The results are shown in Tables~\ref{tab:stat} and \ref{tab:stato1} respectively for zero-order and first-order TSK classifiers, where the statistically significant ones are marked in bold.

\begin{table}[h]\centering \setlength{\tabcolsep}{0.4mm}
\caption{$p$-values of non-parametric multiple comparisons of the RCAs and BCAs on the zero-order algorithms.}\label{tab:stat}
\begin{tabular}{c|lccccc}\toprule
                     &           & FCM-LSE         & EWFCM-LSE       & ESSC-LSE        & SFP             & SESSC           \\\midrule
\multirow{5}{*}{RCA} & EWFCM-LSE & \textbf{0.0003} &                 &                 &                 &                 \\
                     & ESSC-LSE  & \textbf{0.0000} & 0.2927          &                 &                 &                 \\
                     & SFP       & 0.1621          & \textbf{0.0061} & \textbf{0.0016} &                 &                 \\
                     & SESSC     & \textbf{0.0036} & 0.2054          & 0.0978          & 0.0477          &                 \\
                     & SESSC-LSE & \textbf{0.0000} & \textbf{0.0029} & \textbf{0.0105} & \textbf{0.0000} & \textbf{0.0002} \\\midrule
\multirow{5}{*}{BCA} & EWFCM-LSE & \textbf{0.0000} &                 &                 &                 &                 \\
                     & ESSC-LSE  & \textbf{0.0000} & 0.3133          &                 &                 &                 \\
                     & SFP       & 0.1406          & \textbf{0.0024} & \textbf{0.0005} &                 &                 \\
                     & SESSC     & \textbf{0.0000} & 0.1372          & 0.2624          & \textbf{0.0000} &                 \\
                     & SESSC-LSE & \textbf{0.0000} & \textbf{0.0004} & \textbf{0.0019} & \textbf{0.0000} & \textbf{0.0124}\\ \bottomrule
\end{tabular}
\end{table}

\begin{table}[h]\centering
\caption{$p$-values of non-parametric multiple comparisons of the RCAs and BCAs on the first-order algorithms.}\label{tab:stato1}
\begin{tabular}{c|cccc}\toprule
                     &           & FCM-LSE & EWFCM-LSE & ESSC-LSE \\\midrule
\multirow{3}{*}{RCA} & EWFCM-LSE & 0.4934  &           &          \\
                     & ESSC-LSE  & 0.8124  & 0.5079    &          \\
                     & SESSC-LSE & 0.6060  & 0.4405    & 0.4738   \\\midrule
\multirow{3}{*}{BCA} & EWFCM-LSE & 0.1286  &           &          \\
                     & ESSC-LSE  & 0.1319  & 0.5112    &          \\
                     & SESSC-LSE & 0.2636  & 0.6387    & 0.4998  \\\bottomrule
\end{tabular}
\end{table}

Table~\ref{tab:stat} shows that our proposed zero-order SESSC-LSE significantly outperformed all other approaches. In addition, all SSC-based clustering algorithms, except SFP, significantly outperformed FCM-LSE, indicating that the SSC-based clustering algorithms are more suitable for initializing TSK fuzzy classifiers on high-dimensional datasets.

Table~\ref{tab:stato1} shows that different clustering approaches do not affect the performance of first-order TSK fuzzy classifiers significantly.

\subsection{Performances versus the Number of Rules (Clusters)}\label{subsec:diffc}

\begin{figure}[!h]\centering
\includegraphics[clip, width=0.9\columnwidth]{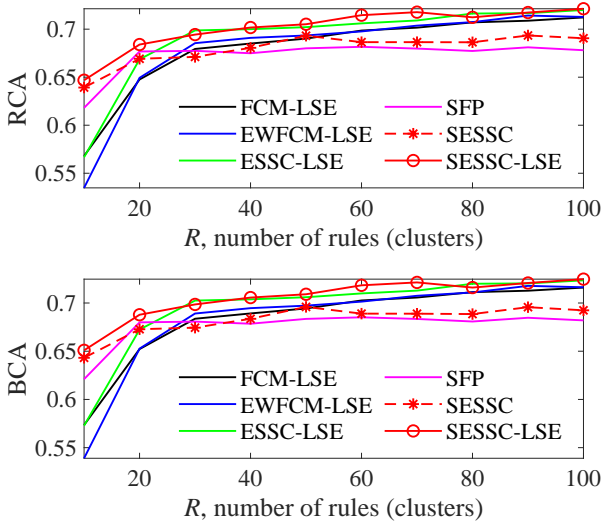}
	\caption{RCAs and BCAs of the six zero-order algorithms with different number of clusters.}\label{fig:diffc}
\end{figure}

Since the performance of a TSK fuzzy classifier varies with $R$, the number of rules, we changed $R$ in  the six zero-order algorithms from 10 to 100, while keeping all other parameters at their optimal values ($\gamma=10$, $\eta=0.1$, $\beta=1$, $\alpha=0.01$ and $h=100$ for SESSC and other TSK fuzzy classifiers; $\gamma'=0.9$, $\alpha'=0.45$ and $\lambda'=0.05$ for SFP) on the Vehicle dataset to study the performances of the five algorithms. The results are shown in Fig.~\ref{fig:diffc}. On average, SESSC-LSE always performed the best. SESSC and SESSC-LSE performed much better than other unsupervised clustering algorithms when $R$ was small.

\subsection{Convergence of SESSC}

\begin{figure}[!h]\centering
	\includegraphics[width=0.9\columnwidth,clip]{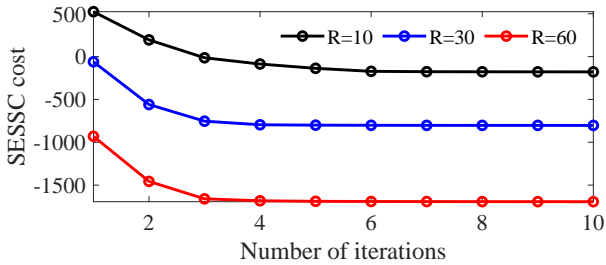}
	\caption{Convergence of the proposed SESSC on Vehicle dataset with different $R$, the number of clusters.} \label{fig:conv}
\end{figure}

We analyzed the convergence of our proposed SESSC algorithm on the Vehicle dataset with different number of clusters. The median value of the cost function from different runs are shown in Fig.~\ref{fig:conv}. SESSC converged quickly after a small number of iterations, e.g., four, which is desirable in practice.

\subsection{Parameter Sensitivity Analysis}

\begin{figure}[!h]\centering
	\subfigure[]{\includegraphics[width=0.48\columnwidth,clip]{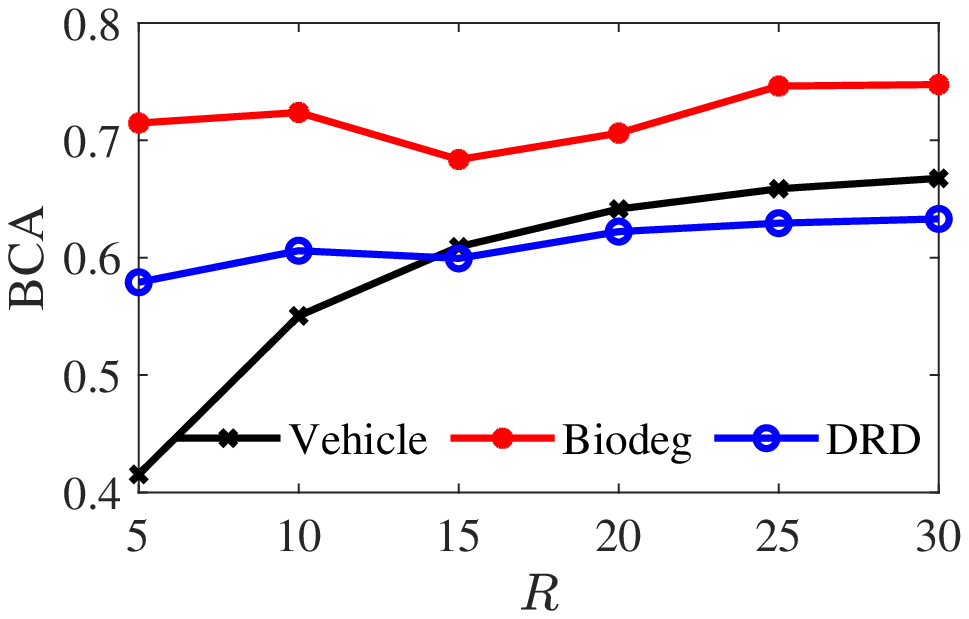}}
	\subfigure[]{\includegraphics[width=0.48\columnwidth,clip]{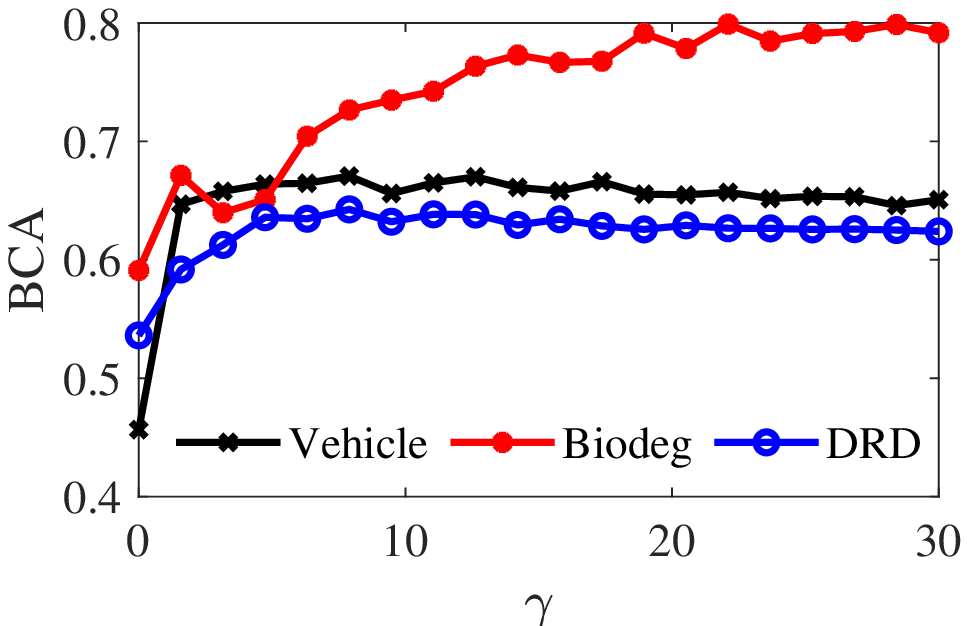}}
	\subfigure[]{\includegraphics[width=0.48\columnwidth,clip]{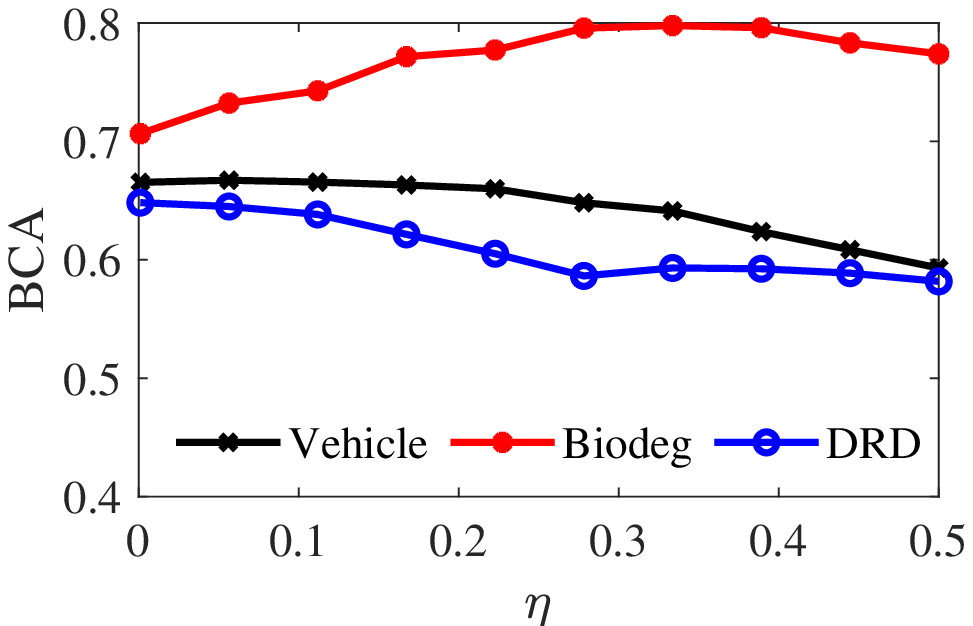}}
	\subfigure[]{\includegraphics[width=0.48\columnwidth,clip]{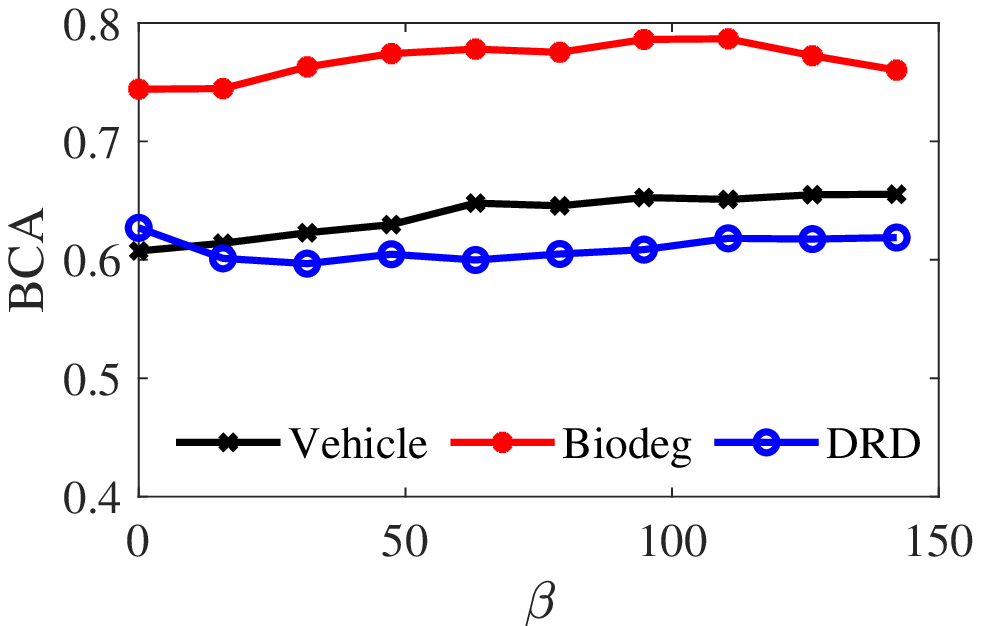}}
	\caption{BCAs with different (a) $R$, (b) $\gamma$, (c) $\eta$ and (d) $\beta$. } \label{fig:param}
\end{figure}

The proposed SESSC algorithm has four parameters, $R$, $\gamma$, $\eta$ and $\beta$. We also tested how they affected the classification performance on three datasets (Vehicle, Biodeg and DRD), by fixing three of them at their default values ($R=30$, $\gamma=10$, $\eta=0.1$ and $\beta=1$) and varying the remaining one. The results are shown in Fig.~\ref{fig:param}. Generally, the performance is stable over a wide parameter range, which is desirable.

\section{Conclusions}\label{sec:con}

This paper has proposed SESSC, a supervised FCM based clustering algorithm to generate the antecedent parameters of TSK fuzzy classifiers. Unlike traditional unsupervised clustering algorithm, SESSC utilizes the label information for better discriminability among different clusters. SESSC is then combined with LSE to further generate the consequent parameters of a TSK fuzzy classifier. Experiments on nine UCI datasets shown that SESSC-LSE outperformed traditional unsupervised FCM based initializations, especially when the number of rules is small.


\end{document}